\documentclass[conference]{IEEEtran}
\usepackage{amsmath,amssymb,amsfonts,mathrsfs,graphicx,hhline,subcaption}
\usepackage[ruled,vlined,linesnumbered]{algorithm2e}
\usepackage[margin=0.7in]{geometry}
\usepackage{booktabs} % For formal tables
\usepackage{CJK}
\usepackage{enumitem,color}

\begin{document}

\title{\emph{EdgeMap}: CrowdSourcing High Definition Map in Automotive Edge Computing\vspace{-0.1in}}

\author{\IEEEauthorblockN{Qiang Liu \\ University of Nebraska-Lincoln \\ qiang.liu@unl.edu\vspace{-0.4in}}
\and
\IEEEauthorblockN{Yuru Zhang \\ Xidian University \\ yuruzhang@stu.xidian.edu.cn\vspace{-0.4in}}
\and
\IEEEauthorblockN{Haoxin Wang \\ University of North Carolina at Charlotte \\ hwang50@uncc.edu\vspace{-0.4in}}
\vspace{-0.3in}}

% \author{\IEEEauthorblockN{Qiang Liu\IEEEauthorrefmark{1}, Yuru Zhang\IEEEauthorrefmark{2}}
% \IEEEauthorblockA{\IEEEauthorrefmark{1} University of Nebraska-Lincoln, NE, United States\\
% \IEEEauthorrefmark{2} Xidian University, Xi’an, China\\
% \IEEEauthorrefmark{1}qiang.liu@unl.edu,
% \IEEEauthorrefmark{2}yuruzhang@stu.xidian.edu.cn\vspace{-0.2in}}}

\maketitle

\begin{abstract}
High definition (HD) map needs to be updated frequently to capture road changes, which is constrained by limited specialized collection vehicles.
To maintain an up-to-date map, we explore crowdsourcing data from connected vehicles.
Updating the map collaboratively is, however, challenging under constrained transmission and computation resources in dynamic networks.
In this paper, we propose \emph{EdgeMap}, a crowdsourcing HD map to minimize the usage of network resources while maintaining the latency requirements.
We design a \emph{DATE} algorithm to adaptively offload vehicular data on a small time scale and reserve network resources on a large time scale, by leveraging the multi-agent deep reinforcement learning and Gaussian process regression.
We evaluate the performance of \emph{EdgeMap} with extensive network simulations in a time-driven end-to-end simulator.
The results show that \emph{EdgeMap} reduces more than 30\% resource usage as compared to state-of-the-art solutions.
\end{abstract}

\begin{IEEEkeywords}
HD Map, Automotive Edge Computing, Multi-Agent Deep Reinforcement Learning
\end{IEEEkeywords}

\vspace{-0.1in}
\section{Introduction}
\label{sec:introduction}
The high definition (HD) map is an essential building block for realizing autonomous driving (AV), as well as advanced driving assistance system (ADAS).
It is developed to provide the most accurate presentation of the roads with extremely high precision, e.g., centimeter-level, which includes various information such as traffic signs and 3D point cloud.
The HD map is built with diverse sensors, e.g., cameras, LiDAR, GPS, IMU, and radars, by using different techniques such as simultaneous localization and mapping (SLAM) algorithms.

To achieve precise localization of vehicles, the HD map needs to incorporate up-to-date road information such as constructions, accidents, and temporary maintenance.
However, this can be challenging because updating the HD map usually requires specialized collection vehicles to traverse each road, collect and process information.
Connected vehicles (CV)~\cite{contreras2017internet}, which connects vehicles with wireless technologies, has been exploited to collaboratively build and update the HD map.
CarMap~\cite{ahmad2020carmap} proposed a lean presentation of SLAM feature maps and substantially decreases the needed bandwidth of map updates.
However, it requires all the computation of the SLAM, e.g., feature extraction and local mapping, to be executed on connected vehicles, which could pose an extra computation burden for vehicles.

Multi-access edge computing (MEC)~\cite{liu2019edge} emerges recently to deploy distributed computation, storage, and networking resources, e.g., edge servers, in close proximity of users, e.g., at the network edge.
By leveraging the capability of MEC, the SLAM computation for building the HD map can be offloaded to the edge server for the acceleration of computations.
Nevertheless, transmitting the massive raw data generated by real-time sensors, e.g., RGB-D images, demands substantial radio resource usage and may overwhelm the radio transmission under a large number of connected vehicles. 
Thus, there is a pushing need to design an HD map that uses minimal resources while maintaining up-to-date information.

In this paper, we propose an HD map, named \emph{EdgeMap}, which crowdsources the data from connected vehicles in automotive edge computing.
We design \emph{EdgeMap} to use the minimum network resources, i.e., radio bandwidth in the RAN and computation resource in the edge server, while satisfying the latency requirements.
We formulate the vehicular offloading and resource reservation problem to seek the optimal offloading decision and resource usage.
We design a novel \emph{DATE} algorithm to decouple the problem into the vehicular offloading and resource reservation subproblem in a small and large time scale, respectively. 
On the one hand, we design an asynchronous multi-agent deep reinforcement learning (aMARL) method to intelligently offload vehicular data based on local network state and partial global state in a distributed manner.
On the other hand, we design a novel learning-based gradient descent method based on Gaussian process regression (GPR) to online learn and update the resource reservation gradually.

The contributions of this paper are summarized as follows:
\begin{itemize}
    \item We design a new HD map, \emph{EdgeMap}, by crowdsourcing data from connected vehicles in automotive edge computing.
    \item We propose a novel \emph{DATE} algorithm to minimize the resource usage of \emph{EdgeMap} while maintaining the latency requirements by managing the vehicular offloading and resource reservation.
    \item We evaluate the performance of \emph{EdgeMap} via extensive network simulations in a time-driven end-to-end simulator. The results show that \emph{EdgeMap} significantly outperforms state-of-the-art solutions. 
\end{itemize}

\begin{figure}[!t]
\centerline{\includegraphics[width=3.4in]{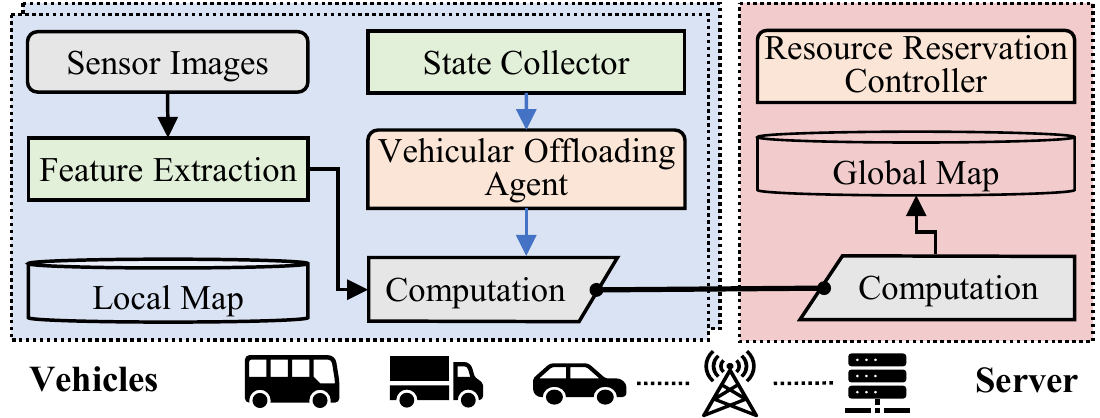}}
\caption{\small \emph{EdgeMap} Overview}
\label{fig:architecture}
\end{figure}

\vspace{-0.1in}
\section{System Overview}
Fig.~\ref{fig:architecture} shows the overview of \emph{EdgeMap}.
In the data plane, the feature extraction module extracts the features from sensor images based on the ORB algorithm~\cite{campos2021orb}.
The features are used for the relocalization of vehicles locally based on the local map and further processed to build the global map based on the SLAM technique~\cite{campos2021orb}.
The SLAM computation can be partially executed in both the vehicle and the edge server, where the split ratio of the computation is determined by the vehicular offloading agent.
The intermediate data generated by the computation in the vehicle is transmitted to the edge server, and the remaining computation is executed in the edge server.
The global map is built and updated after the computation of a connected vehicle is completed.
Then, the new updates in the global map are broadcast to all connected vehicles.

In the control plane, the two main modules, i.e., the vehicular offloading agent in each vehicle and the resource reservation controller, are designed to minimize resource usage while maintaining the latency requirements. 
The vehicular offloading agent is designed based on multi-agent DRL technique to observe local and global network states prepared by the state collector module, e.g., radio channel quality, vehicle mobility, and server workload, and optimize the split ratio of every vehicular offloading in a small time scale (e.g., subseconds).
The resource reservation controller determines the reserved transmission and computation resources from the network operator, which is designed based on Gaussian process regression and runs on a large time scale (e.g., minutes).

\vspace{-0.1in}
\section{System Model and The Problem}
\begin{figure}[!t]
\centerline{\includegraphics[width=3.3in]{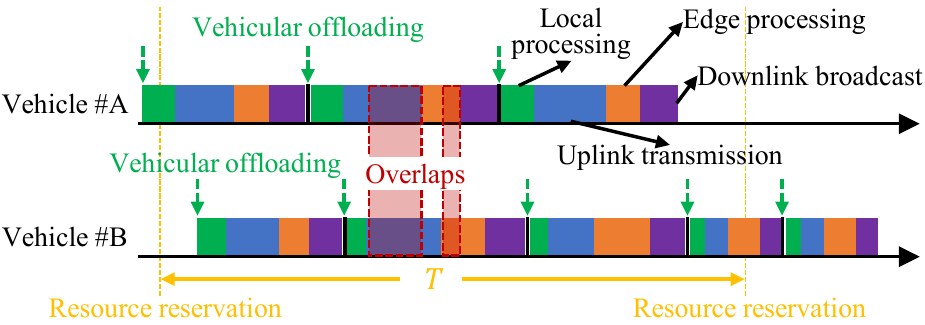}}
\caption{\small An illustration of vehicular offloading }
\label{fig:timeline}
\end{figure}

We consider a mobile network with a cellular base station (BS) and an edge computing server\footnote{This system model can be easily extended to support multiple BSs and edge servers.}.
There are a set of vehicles, denoted as $\mathcal{N}$, that connect to the BS wirelessly.
We illustrate the timeline of \emph{EdgeMap} in Fig.~\ref{fig:timeline}, where two connected vehicles offload the computation to the edge server asynchronously and the resource reservation is configured every $T$ time interval.
Each vehicular offloading is composed of four stages, i.e., local processing, uplink transmission, edge processing and downlink broadcast.
Without loss of generality, we consider the computation can be split\footnote{The continuous computation splitting model can be easily extended to support different schemes such as discrete splitting.} with any ratios ranging from 0 to 1.
The split ratio of $n$th vehicle at the time slot $t$, denoted as $a_n^t \in [0, 1]$, can be determined as the last offloading is completed.
For example, if the split ratio is 0.2, it means there are 20\% computation executed in the vehicle and the remaining 80\% computation are processed in the edge server.
Meanwhile, the intermediate data after the 20\% computation are transmitted to the edge server for remaining computation.
The updates of \emph{EdgeMap}, e.g., new point clouds, are broadcasted to all connected vehicles via downlink broadcast.
Since all connected vehicles are allowed to start their offloadings asynchronously, the vehicular offloadings may overlap with each other, e.g., overlaps occur in both radio transmission and edge processing as shown in Fig.~\ref{fig:timeline}.

To accomplish \emph{EdgeMap}, the service provider requests network resource reservations in multiple technical domains from the network operator\footnote{The isolated resources can be provisioned by the network operator using the network slicing technique~\cite{liu2020edgeslice}.}.
We consider there are three kinds of resources denoted as $\mathcal{M}$, i.e., uplink and downlink radio bandwidth in radio access networks, computing resource in the edge server, i.e., $x_m^T \in [0, 1]$ denotes the reserved $m$th resource at the time $T$.
For the sake of simplicity, we denote the set of split ratios at the time slot $t$ as $\mathcal{A}^t = \{ a_n^t, \forall n \in \mathcal{N}\}$, and the set of resource reservations at the time $T$ as $\mathcal{X}^T = \{ x_m^T, \forall m \in \mathcal{M}\}$.

The \emph{EdgeMap} is designed to provide up-to-date information about the transportation system, e.g., the latest point clouds.
Thus, we define the latency of the $n$th vehicle's offloading started at the time slot $t$ as 
\begin{equation}
    l_n^t = f_n\left( \mathcal{A}^t, \mathcal{X}^T \; |\; \mathcal{S}^t \right), \label{eq:latency}
\end{equation}
where $f_n(\cdot)$ is function with respect to both the split ratios $\mathcal{A}^t$ and the resource reservations $\mathcal{X}^T$ under the network state $\mathcal{S}^t$.
This function is complicated by the asynchronous vehicular offloadings and their overlappings in different stages under high-dim network states. 
As a result, this function cannot be accurately represented by using mathematical expressions\footnote{Although there are several approximated latency models, they did not consider the overlapping scenarios and obtain poor performances (see Sec.~\ref{sec:evaluation}).}.

\textbf{Problem.} We aim to minimize the total resource usage of \emph{EdgeMap} while satisfying the latency requirements.
Therefore, given a time period $\Gamma$, e.g., 1 hour, we formulate the vehicular offloading and resource reservation problem $\mathbb{P}_0$ as
\begin{align}
     \mathbb{P}_0: \min \limits_{ \mathcal{A}^t, \mathcal{X}^T } & \;\;\;\;\; \sum\limits_{t \in \Gamma} {\sum\limits_{m \in \mathcal{M}} {\alpha_m x_m^T}}  \label{eq:org_objective} \\ 
     s.t. &\;\;\;\;\; f_n\left( \mathcal{A}^t, \mathcal{X}^T |\; \mathcal{S}^t\right) \le {L}_{\max}, \forall n, t \label{eq:const_latency} \\
      &\;\;\;\;\; 0 \le a_n^t \le 1, \forall n, t \label{eq:const_x} \\
      &\;\;\;\;\; 0 \le x_m^T \le 1, \forall m, T  \label{eq:const_y}
\end{align}
where $\alpha_m$ is the weight factor for the $m$th resource.
The objective function in Eq.~\ref{eq:org_objective} is defined to minimize the total resource usage and the constraints in Eq.~\ref{eq:const_latency} assure the latency requirements of vehicular offloadings to be satisfied at any time slots.
The constraints in Eq.~\ref{eq:const_x} and Eq.~\ref{eq:const_y} limit the lower and upper bound of the optimization variables, i.e., $\mathcal{A}^t, \mathcal{X}^T, \forall t \in \mathcal{T}$.

\textbf{Challenges.} 
The problem $\mathbb{P}_0$ is difficult to be solved in three aspects.
First, the latency function in Eq.~\ref{eq:latency} is unknown, which makes the problem to be a black-box optimization problem.
As a result, model-based algorithms, e.g., linear programming and convex optimization~\cite{Convex2004Boyd}, can not be used to solve the problem.
Second, the problem is Markovian, where the decision of vehicular offloading affects not only the current latency of vehicles but also the future network state such as server workloads. As a result, the conventional black-box optimization that captures no time-correlated transitions, fails to address this problem.
Third, the coupling optimization variables are heterogeneous in terms of time scales, where the asynchronous vehicular offloading decision of a vehicle can only be made upon its last offloading completion, and the resource reservation happens periodically.

\vspace{-0.1in}
\section{The DATE Algorithm}
\label{sec:algorithm}
In this section, we design a \underline{d}istributed \underline{a}dap\underline{t}ive offloading and r\underline{e}source reservation (DATE) algorithm to effectively solve the problem $\mathbb{P}_0$.
First, we decouple the problem $\mathbb{P}_0$ into two subproblems in different time scales, i.e., a vehicular offloading subproblem to determine the split ratio of vehicular offloadings and a resource reservation subproblem to optimize the reservation of network resources.
Second, we address the vehicular offloading subproblem by proposing an asynchronous multi-agent deep reinforcement learning (aMARL) method based on proximal policy optimization (PPO)~\cite{schulman2017proximal}.
Third, we solve the resource reservation subproblem by proposing a learning-based gradient descent method based on Gaussian process regression (GPR). 

\vspace{-0.1in}
\subsection{Problem Decoupling}
We decouple the problem into two subproblems, which is based on the observation that the resources are reserved in a large time scale (e.g., minutes) and the vehicular offloadings are determined in a small time scale (e.g., subseconds).
In other words, when we optimize the split ratios for vehicular offloadings, the resource reservations are static.

On the one hand, we build the vehicle offloading subproblem under the static resource reservations as
\begin{align}
     \mathbb{P}_1: \min \limits_{ \mathcal{A}^t, \forall t } & \;\;\;\;\;  \sum\limits_{t \in \Gamma} {\sum\limits_{n \in \mathcal{N}} {f_n\left( \mathcal{A}^t \; |\; \mathcal{S}^t, \mathcal{X}^T  \right)}}  \label{prob:offloading}
\end{align}
where we rewrite the original latency function $f_n\left( \mathcal{A}^t, \mathcal{X}^T \; |\; \mathcal{S}^t \right)$ in Eq.~\ref{eq:latency} into $f_n\left( \mathcal{A}^t \; |\; \mathcal{S}^t, \mathcal{X}^T \right)$.

On the other hand, we build the resource reservation subproblem with the static vehicular offloading policy as
\begin{align}
     \mathbb{P}_2: \min \limits_{ \mathcal{X}^T, \forall T } & \;\;\;\;\; \sum\limits_{t \in \Gamma} {\sum\limits_{m \in \mathcal{M}} {\alpha_m x_m^T}}  \\ 
     s.t. &\;\;\;\;\; l_H \le {L}_{\max}, \forall T  \\
      &\;\;\;\;\; 0 \le x_m^T \le 1, \forall m, T
\end{align}
where $l_H = \max\limits_{ 0 \le k \le |\mathcal{N}|}f_k\left( \mathcal{X}^T  \right)$ is the maximum latency among all vehicle offloadings.
Here, we reduce the original latency function $f_n\left( \mathcal{A}^t, \mathcal{X}^T \; |\; \mathcal{S}^t \right)$ in Eq.~\ref{eq:latency} into $f_n\left( \mathcal{X}^T  \right)$, where the split ratios $\mathcal{A}^t$ and states $ \mathcal{S}^t$ in the small time scale are omitted.

\vspace{-0.1in}
\subsection{The Vehicular Offloading Subproblem}
The vehicular offloading subproblem is challenging to solve due to the unknown latency function of vehicle offloadings.
We address this problem by leveraging the multi-agent deep reinforcement learning (MARL) technique that has been increasingly applied to solve networking problems and demonstrated promising performance improvements.

\subsubsection{MARL Basis}
We consider a general multi-agent reinforcement learning setting, where multiple agents $\mathcal{N}$ interact with an environment asynchronously in discrete decision epochs.
At each decision time slot $t$, the agents observe their local states $\mathcal{S}=\{\mathbf{s}_n^{t}, \forall n \in \mathcal{N}\}$, take actions $\mathcal{A}=\{\mathbf{a}_n^t, \forall n \in \mathcal{N}\}$ based on their policies $\pi=\{\pi_n, \forall n \in \mathcal{N}\}$ independently.
These agents receive reward $\mathcal{R}=\{{r}_n^{t}, \forall n \in \mathcal{N}\}$ from the environment, and then the environment transits to the next state based on the actions taken by all agents.
The objective is to find policies $\pi^*=\{\pi_n^*, \forall n \in \mathcal{N}\}$ for all agents to map local states to actions, that maximizes the discounted cumulative reward $R_0 =\sum\nolimits_{t=0}^{T} \gamma^t r_n^{t}$, where $\gamma \in [0, 1)$ is the discounted factor. 

\subsubsection{MARL Design}
To apply the MARL for solving the vehicular offloading problem, we define the state, action space and reward function as follows.

\textbf{State Space.}
We design the state space to incorporate the informative local vehicular status and comprehensive system status.
The first part is $[C_c^{t-1}, C_r^{t-1}, C_s^{t-1}, C_w^{t-1}]$, which are the CPU frequency, RAM, speed, and radio quality of the vehicle at the time slot $t-1$, respectively. These metrics can provide information about hardware specifications and the mobility of the vehicle.
The second part is $[W_s^{t-1}, \mathcal{X}^T]$, which are the server workload at the time slot $t-1$ and current resource reservations, respectively. These metrics indicate the workload of the edge server and available network resources for the system.
Therefore, the state space can be expressed as
\begin{equation}
\label{eq:state}
    \mathbf{s}_t = \left[ C_c^{t-1}, C_r^{t-1}, C_s^{t-1}, C_w^{t-1}, W_s^{t-1}, \mathcal{X}^T \right].
\end{equation}

\textbf{Action Space.}
We design the action space to allow the vehicle to adaptively offload its computation to the edge server.
The action space is expressed as 
\begin{equation}
\label{eq:action}
    \mathbf{a}_t = [ a_{n}^{t} ],
\end{equation}
for the $n$th vehicle at the time slot $t$.

\textbf{Reward.}
We design the reward function to guide the policy update of the agents to minimize the latency of vehicular offloadings as shown in Eq.~\ref{prob:offloading}.
The reward function is expressed as
\begin{align}
\label{eq:reward}
    r_n^t = f_n\left( \mathcal{A}^t \; |\; \mathcal{S}^t \right),
\end{align}

\begin{figure}[!t]
	\centering
	\includegraphics[width=2.0in]{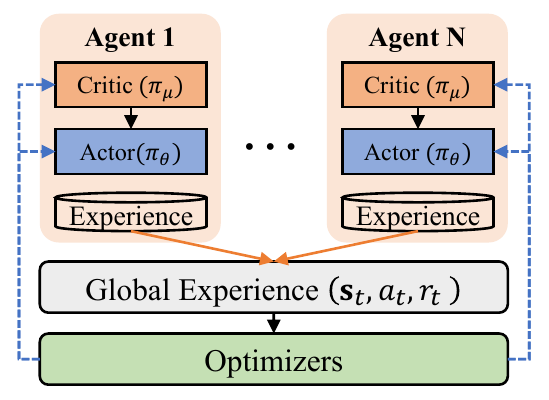}
	\caption{\small The aMARL overview.}
	\label{fig:algorithm}
\end{figure}

The objective is to derive the policies $\mathcal{P}^*=\{\pi_n^*, \forall n \in \mathcal{N}\}$ parameterized by deep neural networks that solve the vehicular offloading problem in Eq.~\ref{prob:offloading}.
To this end, we design all the agents to share the identical policy network $\pi^*$, i.e., parameter sharing~\cite{terry2020revisiting}.
As shown in Fig.~\ref{fig:algorithm}, we create an agent in each vehicle to determine the split ratio of its vehicular offloadings.
Here, we leverage the actor-critic architecture to achieve robust learning performances, i.e., an actor $\pi_\theta$ for taking the action, and a critic $\pi_\mu$ for guiding the update of the actor.
To update the policy network, we centralize the experience from all the agents periodically, calculate the update of the policy network $\pi^*$ globally, and enforce the update for all the agents accordingly.

\subsubsection{MARL Training}
We design the training method of policy networks based on the proximal policy optimization (PPO)~\cite{schulman2017proximal}.
Denote $\pi_\theta$ as the policy with parameters $\theta$, the vanilla policy gradient training method updates the actor policy network at step $k$ according to 
\begin{equation}
    \theta^{k+1} = \theta^k + \eta_0 \cdot \nabla_{\theta} J(\pi_\theta),
\end{equation}
where $\eta_0$ is the step size (i.e., the learning rate), $J(\pi_\theta)$ is the expected finite-horizon return of the policy and 
\begin{equation}
    \nabla_{\theta} J(\pi_\theta) = \mathbb{E}_{\pi_\theta }\left[\sum\limits_{t=0}^{\Gamma}\nabla_{\theta}\log \pi_\theta (a_t|s_t)A^{\pi_\theta} \right] \label{eq:actor_update}
\end{equation}
where $A^{\pi_\theta} = Q^{\pi_\theta}(s_t, a_t) - V^{\pi_\theta}(s_t)$ is the advantage function of the current policy, and $Q^{\pi_\theta}(s_t, a_t)$ and $V^{\pi_\theta}(s_t)$ are the action-value function and the value function, respectively.
However, since this update is unconstrained, a single update step might lead to significant performances difference from the previous policy, and thus can lead to the collapse of policy performance.

We address this issue, based on PPO~\cite{schulman2017proximal}, by updating the actor policy network $\pi_\theta$ at step $k$ via
\begin{equation}
    \theta^{k+1} = \arg \max\limits_{\theta} \mathbb{E}_{\pi_\theta }\left[G(\theta)  \right]. \label{eq:ppo_update}
\end{equation}
Here, the surrogate objective function $G(\theta)$ is expressed as
\begin{align}
    G(\theta) = \min \left[ g_t(\theta)A^{\pi_\theta}, clip\left(g_t(\theta), 1-\epsilon, 1+\epsilon \right) A^{\pi_\theta}\right],
\end{align}
where $g_t(\theta) = {\pi_\theta (a_t|s_t)}/{\pi_{\theta_k} (a_t|s_t)}$ and $\epsilon$ is a hyperparameter to control the difference between the updated policy and the previous policy.
In this way, the update steps with too large policy differences are prevented and the smooth improvement on policy performance is assured.

\vspace{-0.1in}
\subsection{The Resource Reservation Subproblem}
The resource reservation subproblem $\mathbb{P}_2$ is difficult to solve because the latency function $l_H$ is unknown, even if under the static vehicular offloading policies.
We solve this problem in two steps.
First, we use the Gaussian process regression (GPR) to regress the unknown latency function $l_H$ and define expected gradients accordingly.
Second, we design a learning-based gradient descent method to gradually update the resource reservation based on the expected gradients.

\textbf{Gaussian Process Regression.}
Given a resource reservation $\mathcal{A}$, we can obtain the noisy observation $y = l_{H} + \xi$ under the static vehicular offloading policy $\pi^*$, where $\xi$ is the Gaussian noise with a zero mean and $\delta_{noise}^2$ variance.
Denote $\mathcal{X}^{1:B}$ and $y^{1:B}$ as the set of resource reservations and the corresponding observed latencies, respectively. 
The posterior distribution is expressed as
\begin{equation}
	P(l_{H}(\mathcal{X})|\mathcal{D}^{1:B}) \propto P(\mathcal{D}^{1:B}|l_{H}(\mathcal{X}))P(l_{H}(\mathcal{X})). \label{eq:gaussian_posterior}
\end{equation}
where the observations denote $\mathcal{D}^{1:B}=\{\mathcal{X}^{1:B}, l_{H}^{1:B}\}$.
We use the Gaussian process (GP) to model the prior distribution $l_{H}(\mathcal{X})$~\cite{rasmussen2003gaussian}. 
Here, $l_{H}(\mathcal{X})$ can be described as $l_{H}(\mathcal{X}) \sim \mathcal{GP}(\mu(\mathcal{X}), k(\mathcal{X},\mathcal{X}'))$ where $\mu(\mathcal{X})$ and $k(\mathcal{X},\mathcal{X}')=\exp(-\frac{1}{2}||\mathcal{X}-\mathcal{X}'||^2)$ are the mean function and covariance function, respectively.

Given a resource allocation $\tilde{\mathcal{X}}$, the posterior distribution can be derived as
\begin{equation}\label{eq:predict_distribution}
	P(l_{H}(\tilde{\mathcal{X}})|\mathcal{D}^{1:B},\tilde{\mathcal{X}}) \sim \mathcal{N}(\mu(\tilde{\mathcal{X}}),\sigma^2(\tilde{\mathcal{X}})),
\end{equation}
where $\mu(\tilde{\mathcal{X}}) = \mathbf{k}^T[\mathbf{K}+\delta^2 \mathbf{I}]^{-1}y^{1:B}$, and 
\begin{equation}\label{eq:statics_predict_2}
	\sigma^2_{i}(\tilde{\mathcal{X}}) = k(\tilde{\mathcal{X}},\tilde{\mathcal{X}})-\mathbf{k}^T[\mathbf{K}+\delta_{noise}^2 \mathbf{I}]^{-1}\mathbf{k},
\end{equation}
%are the sufficient statistics of expected posterior distribution $P(l_{H}(\mathcal{X}^*)|\mathcal{D}^{1:B},\tilde{\mathcal{X}})$.
where $\mathbf{k}=[k(\tilde{\mathcal{X}}, \mathcal{X}^{1}), k(\tilde{\mathcal{X}}, \mathcal{X}^{2}), \cdots, k(\tilde{\mathcal{X}}, \mathcal{X}^{B})]$ and $\mathbf{K}= [k(\mathcal{X}^{i}, \mathcal{X}^{j})]_{B\times B}, \forall i, j =1,2,...,B$.

\textbf{Expected Gradient.}
Based on the posterior distribution, we define the expected gradient~\cite{liu2019virtualedge} under the resource reservation $\mathcal{X}$ as
\begin{equation}\label{eq:expected_gradient}
	\nabla {l}_H({\mathcal{X}}) = 0.5 * \left( \mu(\mathcal{X} + \tau*\mathbf{I})- \mu(\mathcal{X} - \tau*\mathbf{I}) \right) / \tau,
\end{equation}
where $\mathbf{I}$ and $\tau$ are an identity matrix and small positive constant, respectively. 

Then, we solve the resource reservation subproblem $\mathbb{P}_2$ by using the subgradient descent method~\cite{Convex2004Boyd} based on the expected gradients. The resource reservation at the time interval $T$ is updated according to
\begin{align}
	\mathcal{X}^{T+1} &= clip\left(\mathcal{X}^{T} + \mathbf{\eta}_1 (1 - \lambda \cdot \nabla {l}_H(\mathcal{X}^T)), 0, 1 \right), \label{eq:resource_update} \\
	\lambda^{T+1} &= [\lambda^{T} + \mathbf{\eta}_2( l_H - {L}_{\max})]^+, \label{eq:multiplier_update}
\end{align}
where $\mathbf{\eta}_1, \mathbf{\eta}_2$ are appropriate step sizes, $\lambda$ is the multiplier and $[x]^+ \triangleq \max(x,0)$.

\vspace{-0.1in}
\subsection{The DATE Algorithm }
We summarize the \underline{d}istributed \underline{a}dap\underline{t}ive offloading and r\underline{e}source reservation (DATE) algorithm as follows.
First, we train the policy networks under the varying resource reservations.
As the time to reserve resources, we centralize the transitions of all agents and used them to update the policy networks.
After the training converges, we can obtain the optimal policy network $\pi^*$.
Second, we train the Gaussian process regression model with the optimal policy network.
We use the optimal policy network $\pi^*$ to rollout the vehicular offloading until the time to reserve resources.
The latency $l_H$ is calculated and the resource-latency pair is stored in the GPR training dataset.
The GPR model is updated as new pairs are added in the training dataset.
The resource reservation and the multiplier are also updated according to Eq.~\ref{eq:resource_update} and Eq.~\ref{eq:multiplier_update}.
As the algorithm completes, we obtain the optimal policy network $\pi^*$ and the well-trained GPR model.

\vspace{-0.1in}
\section{Performance Evaluation}
\label{sec:evaluation}
In this section, we evaluate the performance of \emph{EdgeMap} with extensive network simulations.

\begin{figure}[t]
	\centering
	\includegraphics[width=3.3in]{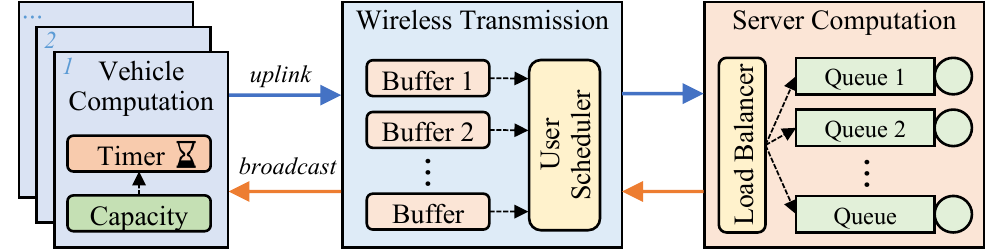}
	\caption{\small The design of network simulator.}
	\label{fig:simulator}
\end{figure}

\textbf{End-to-End Simulator.}
We build a time-driven network simulator, which is composed of multiple vehicle computation modules, a wireless transmission module, and a server computation module as depicted in Fig.~\ref{fig:simulator}.
For each simulation time, e.g., 1ms, the tasks in either vehicle and server computation modules decrease their remaining time.
Meanwhile, the tasks in either uplink and broadcast transmission modules update the remaining buffer size.
In particular, when the vehicular offloading decision is determined, a task is created on this vehicle's computation module.
The task describes the remaining vehicle computation time, data size of uplink and broadcast transmission packet, and server computation time, where these data are sampled from experimental measurements.
The simulation of server computation is based on multiple parallel FIFO (First-In-First-Out) service queues, where a default \emph{min-load} scheduling scheme is applied to choose the queue for serving the incoming task.
Besides, the higher computation capacity of the server reduces the computation time of the task in each service queue.

The wireless transmission module is developed based on an open-source 5G simulator~\cite{oughton2019open}, where we use the urban micro (UMi - Street Canyon) channel model recommended in ETSI TR 138.901~\cite{esti_tr_138_901}.
The maximum wireless bandwidth for uplink and downlink channels are 5MHz and the bandwidth is equally shared by all the vehicles if they have traffic to transmit (non-empty buffer).
Thus, the transmission of tasks are simulated by calculating their wireless data rates\footnote{The mobility of vehicles and the location of the base station are obtained from a trace, which is collected from an intersection scenario built in Unity3d~\cite{liu2021livemap}.} and decreasing their remaining uplink/downlink data sizes.
A task is sent to the next simulation module only if it is completed in the last module, e.g., zero remaining computation time or transmission data size.
Each vehicle starts the next vehicular offloading as the last offloading completes.

\textbf{DRL agents.}
We implement multiple DRL agents by using PyTorch 1.9.1.
We use 3-layer fully-connected neural network in both actor and critic networks, where there are 128 neurons in both layers with Leaky Recifier~\cite{goodfellow2016deep} activation functions, and adopt $sigmoid$~\cite{goodfellow2016deep} as the activation function at the output layer of the actor.
On training the DRL agents, we use the following hyper-parameters.
The learning rates of actor and critic networks are 1e-4 and 3e-4, respectively. The maximum episode length is 100, an epoch includes 4000 transitions, and $\epsilon=0.2$ and $\eta_1=0.02, \eta_2=0.02$.
We add the decaying Gaussian noise on actions $\mathbf{a}_t$ during the training phase for balancing the exploitation and exploration.

We use \emph{scikit-learn} v1.0.0~\cite{scikit-learn} to build the Gaussian process regression (GPR) model.
The resource reservation can be configured every 100 vehicular offloadings.

\begin{figure*}[!t] % cannot have space 
\captionsetup{justification=centering}
  \begin{minipage}[t]{0.325\textwidth}
    \centering
    \includegraphics[width=2.4in, height=1.0in]{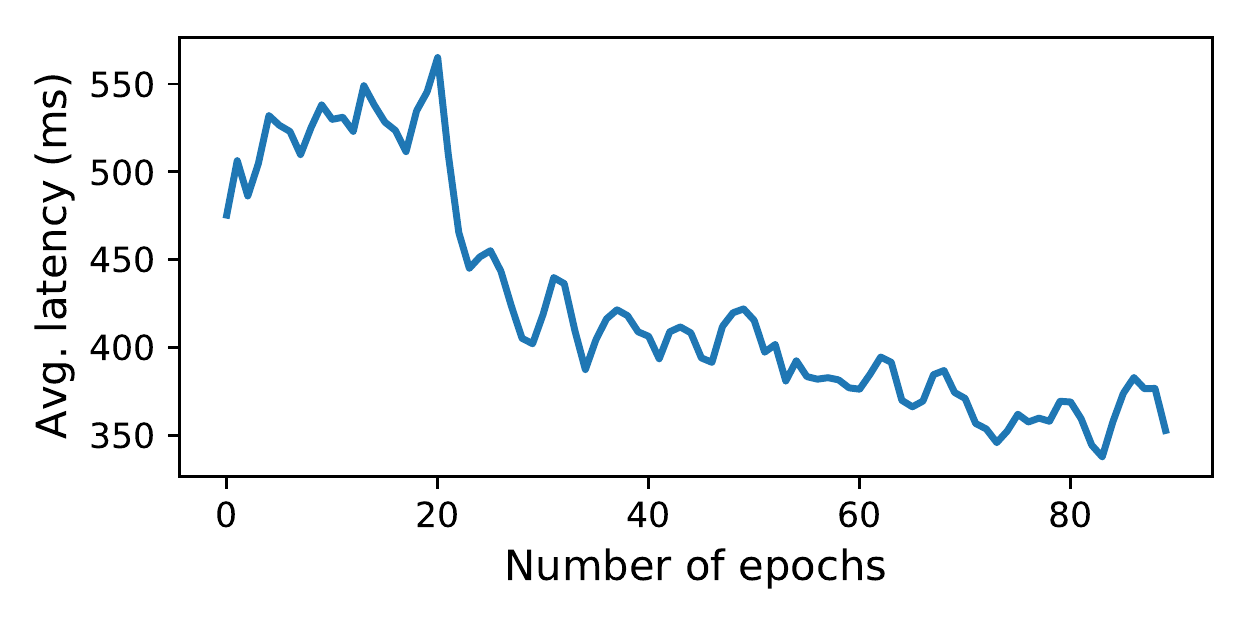}
    \captionof{figure}{\small Training of aMARL agents}
    \label{fig:result_main_training}
  \end{minipage}
  \hfill
  \begin{minipage}[t]{0.325\textwidth}
    \centering
    \includegraphics[width=2.4in, height=1.0in]{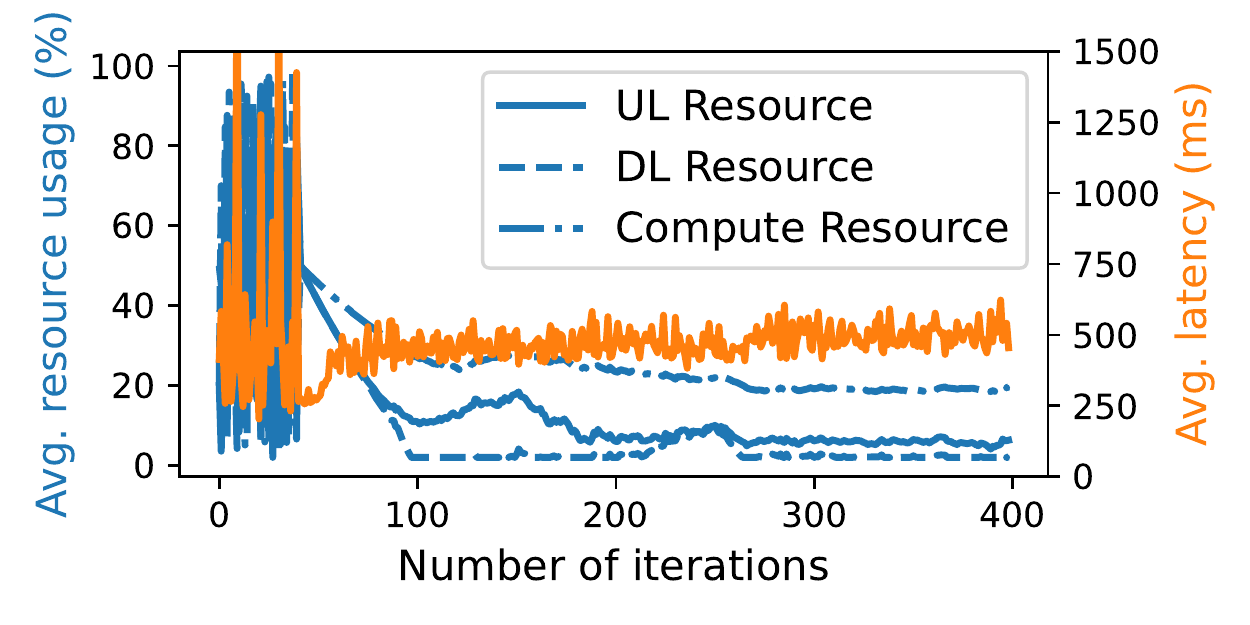}
    \captionof{figure}{\small Convergence of \emph{EdgeMap}}
    \label{fig:results_overall_convergence}
  \end{minipage}
  \hfill
  \begin{minipage}[t]{0.325\textwidth}
    \centering
    \includegraphics[width=2.4in, height=1.0in]{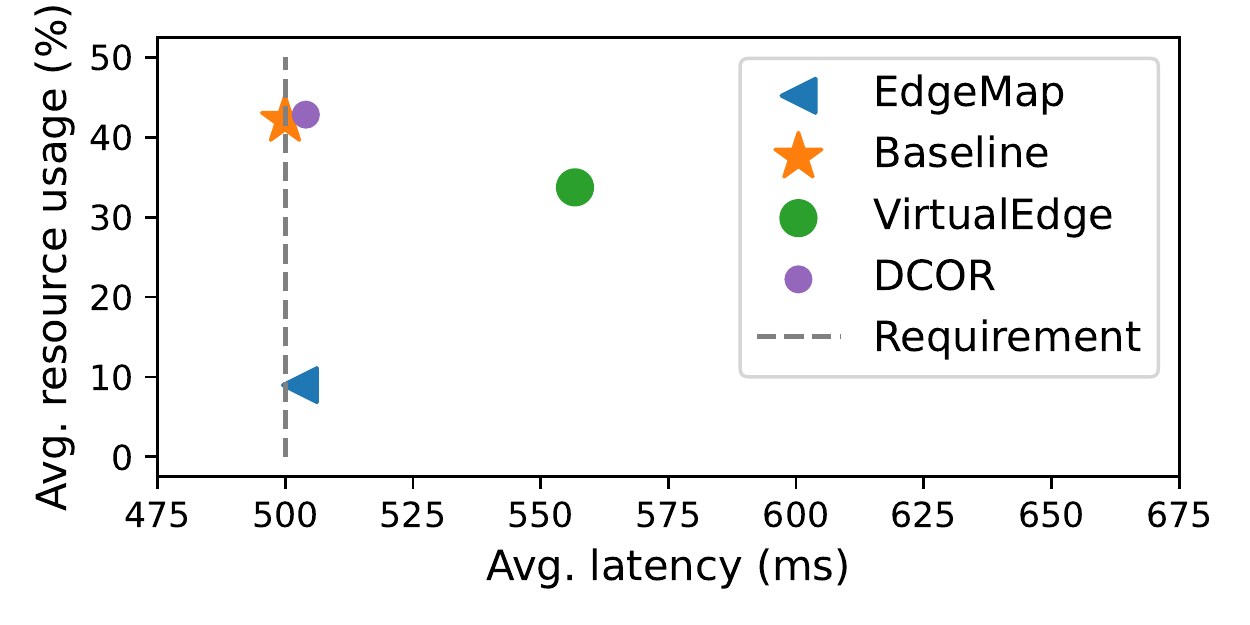}
    \captionof{figure}{\small Performance of algorithms}
    \label{fig:results_algorithms}
  \end{minipage}
  \hfill
\end{figure*}

\begin{figure*}[!t] % cannot have space 
\captionsetup{justification=centering}
  \begin{minipage}[t]{0.325\textwidth}
    \centering
    \includegraphics[width=2.4in, height=1.0in]{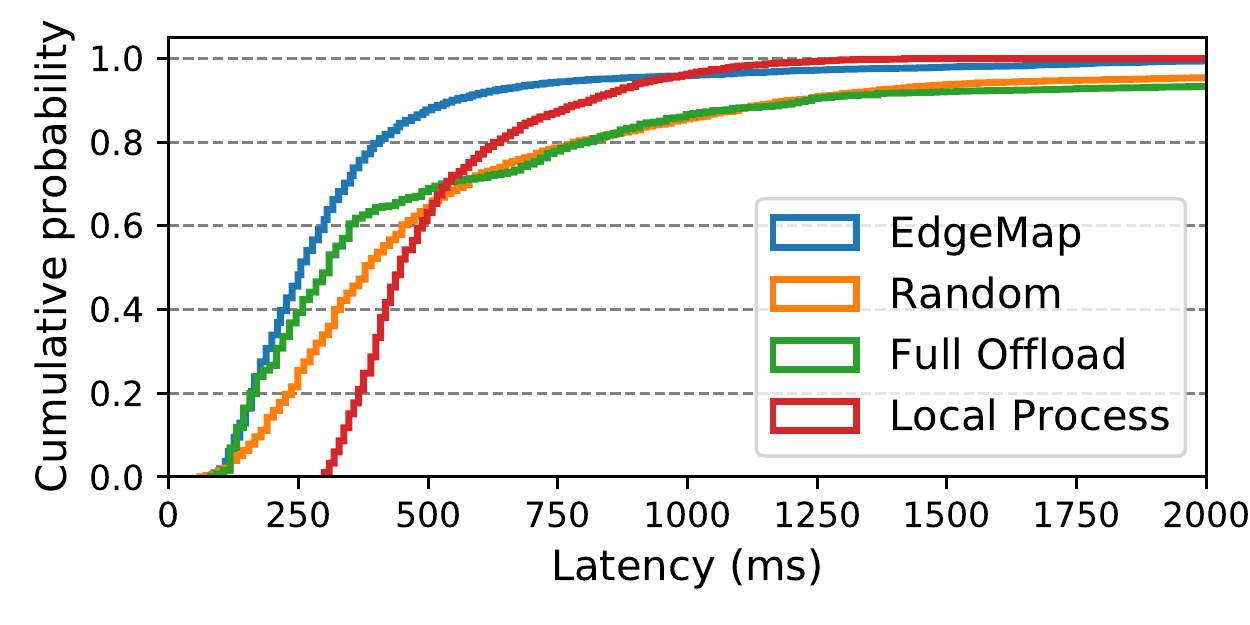}
    \captionof{figure}{\small Offloading latency of methods}
    \label{fig:results_offload_latency}
  \end{minipage}
  \hfill
  \begin{minipage}[t]{0.325\textwidth}
    \centering
    \includegraphics[width=2.4in, height=1.0in]{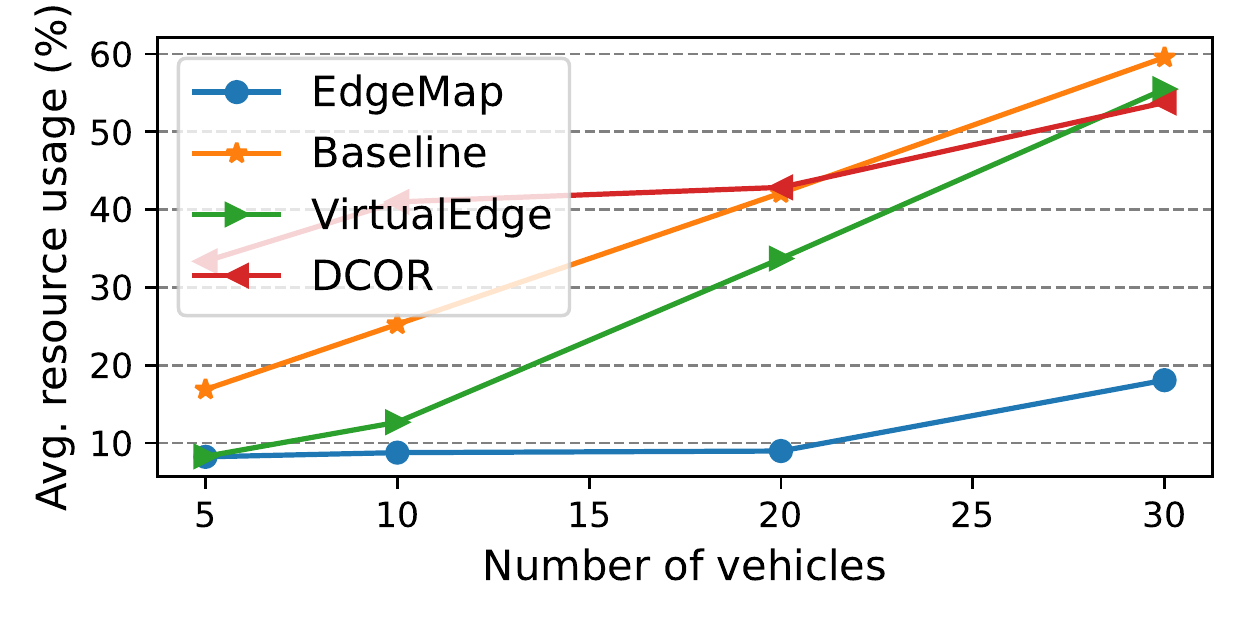}
    \captionof{figure}{\small Scalability of \emph{EdgeMap} }
    \label{fig:results_num_of_vehicles}
  \end{minipage}
  \hfill
  \begin{minipage}[t]{0.325\textwidth}
    \centering
    \includegraphics[width=2.4in, height=1.0in]{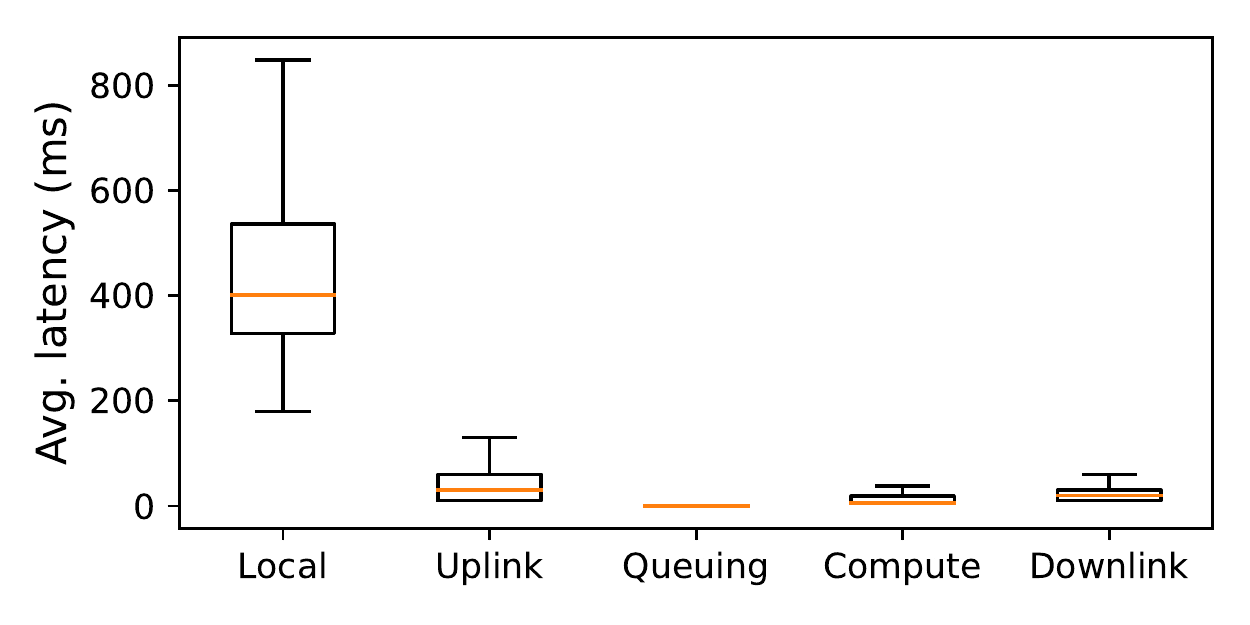}
    \captionof{figure}{\small Latency in different stages}
    \label{fig:result_latency_detail}
  \end{minipage}
  \hfill
\end{figure*}

\textbf{Parameters.}
To obtain the transmission and computation demands of vehicular offloadings, we conduct the experiments with ORB-SLAM3~\cite{campos2021orb} on the Mono mode.
We use the EuRoC dataset with the scenarios of \emph{Machine Hall} 01-05. %~\cite{burri2016euroc}
We measure the data size of images, which have 353.5KB mean and 22.7KB std.
We profile the ORB-SLAM3 computation latency with both a Desktop (AMD Ryzen 3600 3.8GHz, 32GB RAM) as the edge server and a Laptop (Intel i7-6500U 2.5GHz, 8GB RAM) as a vehicle.
The computation latency in the Desktop and the Laptop are 286.54ms mean and 68.89ms std, and 609.27ms mean and 165.44ms std, respectively.
Here the computation latency includes the tracking and local mapping, and excludes the global mapping.
To determine the intermediate data size and computation under different split ratios, we linearly interpolate the data size for radio transmission and computation latency in the edge server.
The weight factors for all kinds of resources are $\alpha_m = 1, \forall m$.

\textbf{Comparison.}
We compare \emph{EdgeMap} with the following algorithms:
\begin{itemize}
    \item \emph{Baseline}: The \emph{Baseline} offloads all the images to the edge server, i.e., the split ratio is always 0. The resource reservation is determined by decreasing from the maximum resource usage until the latency requirements are met.
    \item \emph{DCOR}: The \emph{DCOR}~\cite{zhao2019computation} builds the mathematical models of the latency function in Eq.~\ref{eq:latency} and solves the vehicular offloading and resource reservation sub-problem alternatively. The problem of minimizing the overall resource usage is solved by using the \emph{CVXPY} tool.%~\cite{agrawal2018rewriting}
    \item \emph{VirtualEdge}: The \emph{VirtualEdge}~\cite{liu2019virtualedge} offloads all the images to the edge server, uses GPR to regress the maximum latency function $l_H$ and updates the resource reservation with a proximal gradient descent method.
\end{itemize}

\textbf{Convergence.}
Fig.~\ref{fig:result_main_training} shows the average latency of vehicular offloadings achieved by \emph{EdgeMap} during the training phase of aMARL agents.
It can be observed that the aMARL agents gradually learn and reduce the latency of vehicular offloadings in 80 epochs.
As compared to the initial performance at the epoch 0, i.e., 475.4ms average latency, the final trained aMARL agents achieve 352.8ms average latency, i.e., 25.8\% latency reduction.
This validates the effectiveness of the \emph{DATE} algorithm on training multiple asynchronous agents simultaneously.
Fig.~\ref{fig:results_overall_convergence} shows the overall convergence of \emph{EdgeMap}, where the resource reservations are updated with the trained aMARL agents.
In the first 40 iterations, we allow random exploration for initializing the GPR model.
After that, the latency performance converges to the latency requirements (i.e., 500ms), along with the usage decrease on different resources.
Eventually, \emph{EdgeMap} uses 6.2\% uplink radio bandwidth (0.31 MHz), 2\% downlink radio bandwidth (0.1 MHz) and 19.4\% computation resource (1.94 server capacity).
This result justifies \emph{EdgeMap} can achieve a significant reduction of resource usage while maintaining the given latency requirements.

\textbf{Performance.}
Fig.~\ref{fig:results_algorithms} shows both latency and resource performance of different algorithms.
It can be seen that \emph{Baseline} and \emph{DCOR} have similar performance, which satisfy the latency requirement and use nearly 42\% resources on average.
\emph{EdgeMap} uses 9.2\% average resource usage and obtains the average latency of 502.8 ms, in other words, it can reduce nearly 33\% average resource usage as compared to \emph{Baseline} and \emph{DCOR}.
Fig.~\ref{fig:results_offload_latency} shows the cumulative probability of vehicular offloading latency under different methods.
This result is obtained by randomly generating resource reservations and measuring the latency of each vehicular offloading.
It can be observed that \emph{EdgeMap} achieves the lowest latency (338.5ms mean) and the local process method gets the highest latency (716.0ms mean).
This result validates that \emph{EdgeMap} can intelligently optimize the adaptive vehicular offloading under different resource reservations.

\textbf{Scalability.}
Fig.~\ref{fig:results_num_of_vehicles} shows the average resource usage of algorithms under different number of vehicles in the network.
It can be seen that \emph{EdgeMap} can always obtain the lowest average resource usage as compared to other algorithms, and the performance gap becomes larger as the number of vehicles increases.
This result indicates \emph{EdgeMap} can scale to manage vehicular offloading and resource reservation in large-scale networks with more vehicles.
Fig.~\ref{fig:result_latency_detail} shows the latency of vehicular offloading obtained by \emph{EdgeMap} in different stages, i.e., local processing, uplink transmission, queuing and computing in the edge server, and the downlink broadcast.
It can be seen that vehicular offloadings spend a major of time in the local processing stage on vehicles.
This is because \emph{EdgeMap} aims to minimize the resource usage, and the computing capacity on vehicles are exploited as much as possible to decrease the transmission data size and remaining computation at the edge server.
For example, \emph{EdgeMap} may need minimal resources if the local processing alone can satisfy the latency requirements.

\vspace{-0.1in}
\section{Related Work}
This work relates to the computation offloading and resource allocation in edge computing, which has been extensively studied with both model-based and model-free approaches.
The model-based approaches~\cite{wang2017computation, zhao2019computation, zhang2018joint} formulate the problem with mathematical expressions and derive the algorithms based on linear and non-linear, and convex optimization. For example, Zhao \emph{et. al} proposed a distributed computation offloading and resource allocation (DCOR) algorithm to alternatively solve the subproblem of computation offloading and resource allocation.
However, these approaches fail to accurately represent the problem due to the complex network dynamics, e.g., overlapping of computation offloadings.
The mode-free approaches~\cite{li2018deep, liu2019deep, chen2021drl, nath2020multi} exploited the advanced machine learning techniques, e.g., DRL, to handle the high-dim network state and dynamics.
For instance, Chen \emph{et. al} proposed a temporal attentional deterministic policy gradient (TADPG) to improve task completion time and the average energy consumption by optimizing the partial task offloading.
However, these approaches overlook the different time scales of computation offloading (in subseconds) and resource reservation (in minutes), and thus cannot be implemented in real operational networks.

\vspace{-0.1in}
\section{Conclusion}
In this paper, we have presented \emph{EdgeMap}, a crowdsourcing HD map in automotive edge computing. We have formulated the vehicular offloading and resource reservation problem and proposed the \emph{DATE} algorithm to solve it effectively. The performance of \emph{EdgeMap} has been validated with extensive network simulations. The results have shown that \emph{EdgeMap} significantly outperforms the state-of-the-art solutions in terms of performance and scalability.

\vspace{-0.1in}
\bibliographystyle{IEEEtran}
\bibliography{ref/reference}

\end{document}